\documentclass[final]{siamart1116}

\usepackage{lipsum}
\usepackage{amsfonts}
\usepackage{graphicx}
\usepackage{epstopdf}
\usepackage{algorithmic}
\ifpdf
  \DeclareGraphicsExtensions{.eps,.pdf,.png,.jpg}
\else
  \DeclareGraphicsExtensions{.eps}
\fi

\usepackage{xspace}

\DeclareMathOperator*{\argmin}{arg\,min}

\newcommand{\Cr}{\ensuremath{\mathcal{C}^{r}}\xspace}
\newcommand{\Man}{\ensuremath{\mathcal{M}}\xspace}
\newcommand{\MyExp}[2]{\ensuremath{\mathbf{Exp}_{#1}(#2)}\xspace}
\newcommand{\Gnk}[2]{\ensuremath{\mathtt{G}(#1,#2)}\xspace}
\newcommand{\Rn}[1]{\ensuremath{\mathbb{R}^{#1}}\xspace}
\newcommand{\TMan}[1]{\ensuremath{\mathcal{T}_{#1}\mathcal{M}}\xspace}
\newcommand{\vecsub}[2]{\ensuremath{\mathbf{#1}_{#2}}\xspace}

\numberwithin{theorem}{section}

\newcommand{\TheTitle}{A Quasi-isometric Embedding Algorithm} 
\newcommand{\TheAuthors}{David W. Dreisigmeyer}

\headers{\TheTitle}{\TheAuthors}

\title{{\TheTitle}\thanks{
Any opinions and conclusions expressed herein are those of the author and do not necessarily represent the views of the U.S. Census Bureau.
The research in this paper does not use any confidential Census Bureau information.
}}

\author{
	David W. Dreisigmeyer\thanks{United States Census Bureau, 
Center for Economic Studies, 
Suitland, MD and
Department of Electrical and Computer Engineering, 
Colorado State University, 
Fort Collins, CO
    (\email{david.wayne.dreisigmeyer@census.gov})}
}

\usepackage{amsopn}

\ifpdf
\hypersetup{
  pdftitle={\TheTitle},
  pdfauthor={\TheAuthors}
}
\fi

\begin{document}

\maketitle

\begin{abstract}
  The Whitney embedding theorem gives an upper bound on the smallest embedding dimension of a manifold.
  If a data set lies on a manifold, a random projection into this reduced dimension will retain the manifold structure.
  Here we present an algorithm to find a projection that distorts the data as little as possible.
\end{abstract}

\begin{keywords}
  dimensionality reduction, Whitney embedding theorem
\end{keywords}

\begin{AMS}
  15-04, 57-04
\end{AMS}

\section{\label{sec:introduction}Introduction}
Reducing the ambient dimensionality of a data set is a common preprocessing step in data mining.
Often the data can be taken to lie on some high-dimensional manifold.
Typical examples of this would be images or stationary time series.
A central result in differential topology is Whitney's embedding theorem which states
\begin{theorem}[Whitney's embedding theorem \cite{book:hirsch-1997}]
\label{th:whitney}
Let \Man be a (compact Hausdorff) \Cr $n$-dimensional manifold, $2 \leq r \leq \infty$. 
Then there is a \Cr embedding of \Man in \Rn{2n+1}.
\end{theorem}
The method of proof for Theorem~\ref{th:whitney} is, roughly speaking, to find a $(2n+1)$-plane in \Rn{m} such that none of the secant and tangent vectors associated with \Man are completely collapsed when \Man is projected onto this hyperplane.
Almost surely any $(2n+1)$-plane satisfies this condition.
The hyperplane is a point $p$ on the Grassmannian $\Gnk{m}{2n+1}$, the manifold of all $(2n + 1)$-dimensional subspaces of \Rn{m}.  
Then $p \in \Gnk{m}{2n+1}$ contains the low-dimensional embedding of our manifold \Man via the projection $p^{T} \Man \subset \Rn{2n+1}$.
Numerically a random point $p \in \Gnk{m}{2n+1}$ can be chosen, which means that for manifold-valued data on a $n$-dimensional manifold a random projection into \Rn{2n+1} typically gives an embedding \cite{art:broomhead-2000, art:broomhead-2005}.

An important idea is to make this embedding cause as little distortion as possible. 
By this we mean: What $p \in \Gnk{m}{2n+1}$ will minimize the maximum collapse of the worst tangent vector projection?
In this way we can keep the low-dimensional embedding from almost self-intersecting as much as possible.

The current paper is organized as follows.
In Section~\ref{sec:algorithm} we present the dimensionality reduction algorithm.
Section~\ref{sec:examples} looks at an example of reducing the dimensionality of the MNIST dataset of handwritten digits \cite{art:lecun-1998}.
A discussion follows in Section~\ref{sec:discussion}.

\section{\label{sec:algorithm}The Embedding Algorithm}
In practice, we will only have some set $\mathcal{P} = \left\lbrace \vecsub{x}{i} | \vecsub{x}{i} \in \Man \subset \Rn{m} \right\rbrace$ of sample points from our manifold \Man. 
We can then form the set of unit length secants $\Sigma$ that we have available to us, where
\begin{subequations}
\begin{eqnarray}
\label{eq:secants}
\Sigma 
	& = & 
		\left\lbrace
        	\left.
                \frac
                    {\vecsub{x}{i} - \vecsub{x}{j}}
                    {\| \vecsub{x}{i} - \vecsub{x}{j} \|_{2}}
			\ \right\rvert \ 
            	\vecsub{x}{i}, \vecsub{x}{j} \in \mathcal{P} \mbox{ and } i \neq j
        \right\rbrace
        \mbox{.}
\end{eqnarray}
For a given projection $p \in \Gnk{m}{k}$, $k = 2n+1$, the distortion of a secant $\sigma \in \Sigma$ is defined as $| 1 - \| p^{T} \sigma \|_{2}^{2} |$.
The distortion associated with any $p$ is
\begin{eqnarray}
\label{eq:distortion}
	D_{\Sigma}(p) 
    	& = &
    		\max_{\sigma \in \Sigma} 
            	\left\lvert 
                	1 - \| p^{T} \sigma \|_{2}^{2}
			\right\rvert
        \mbox{.}
\end{eqnarray}

We are looking for a point $p \in \Gnk{m}{k}$ that minimizes $D_{\Sigma}(p)$.
This gives the optimization problem
\begin{eqnarray}
\label{eq:opt-prob}
	\widehat{p} 
    	& = & 
    		\argmin_{p \in \Gnk{m}{k}} D_{\Sigma}(p)
    	\mbox{.}
\end{eqnarray}
\end{subequations}
The function $D_{\Sigma}(p)$ is Lipschitz but not differentiable.
A derivative-free optimization algorithm is appropriate.
Direct search methods over manifolds such as the Grassmannian have been developed in \cite{misc:dreisigmeyer-2007,misc:dreisigmeyer-2017} following on the work in \cite{art:edelman-1999}.

The general idea of doing a direct search over an $n$-dimensional (Riemannian) manifold \Man is to work in the tangent space of a point $p \in \Man$.
The tangent space \TMan{p} is a vector space with an inner product and, therefore, is as easy to work with as \Rn{n}.
In the remainder we will let $\Man \equiv \Gnk{m}{k}$.
Properties of Grassmannians are developed in \cite{art:edelman-1999}.

\begin{subequations}
\label{eq:Gnk-opt-info}
For a given $p \in \Man$ the tangent space is 
\begin{eqnarray}
\label{eq:TGnk}
	\TMan{p} 
		& = & 
    		\left\lbrace \omega \ | \ p^{T} \omega = 0 \right\rbrace
	\mbox{.}
\end{eqnarray}
The dimensionality of \Man, and therefore \TMan{p}, is $k(m-k)$.
So it follows that $\TMan{p} \sim \Rn{k(m-k)}$.
The inner product between $\omega_{1}, \omega_{2} \in \TMan{p}$ is 
\begin{eqnarray}
\label{eq:TGnk-metric}
	h(\omega_{1}, \omega_{2}) 
    	& = & \mathrm{Tr}(\omega_{1}^{T} \omega_{2})
	\mbox{,}
\end{eqnarray}
with $\mathrm{Tr}(\cdot)$ the matrix trace operation.

The only additional step required for optimizing over \Man is the need to map \TMan{p} onto \Man.
This is done by the exponential map $\mathbf{Exp}_{p} : \TMan{p} \rightarrow \Man$.
In the current situation the exponential map has a convenient closed-form solution.
For the point $\omega \in \TMan{p}$ let the singular value decomposition be given by $\omega = U \Theta V^{T}$.
Then 
\begin{eqnarray}
\label{eq:TGnk-Exp}
\MyExp{p}{\omega} 
	& = & 
    	\left[ p V \cos(\Theta) + U \sin(\Theta) \right] V^{T} \mbox{.}
\end{eqnarray}
\end{subequations}
With this mapping the optimization problem (\ref{eq:opt-prob}) stated over \Man can be restated as a problem over $\TMan{p} \sim \Rn{k(m-k)}$ at a fixed $p \in \Man$:
\begin{eqnarray}
\label{eq:opt-prob-Rn}
	\widehat{\omega} 
    	& = & 
    		\argmin_{\omega \in \TMan{p}} D_{\Sigma} \circ \MyExp{p}{\omega}
    	\mbox{.}
\end{eqnarray}
Solving (\ref{eq:opt-prob-Rn}) gives the quasi-isometric embedding algorithm.
We call this quasi-isometric because, while it contracts every (secant) tangent vector, the contraction of any (secant) tangent is minimized.

\section{\label{sec:examples}Example}
The MNIST database is composed of $28$-by-$28$ images of handwritten digits divided into a training set of 60000 images and a test set of 10000 images \cite{art:lecun-1998}.
Here the training set is further divided into 50000 example images and 10000 validation images\footnote{The Python dataset mnist.pkl.gz is available at \href{http://deeplearning.net/data/mnist/}{http://deeplearning.net/data/mnist/}.}.
Every image is taken to lie on a manifold in \Rn{784} with ten separate manifolds, one per digit.

Each of the image manifolds can separately have its dimensionality reduced by performing the optimization in (\ref{eq:opt-prob-Rn}).
The initial $p \in \Gnk{784}{k}$ is taken as the leading $k$ columns of $U$ in the SVD $\Sigma = USV^{T}$, with $\Sigma$ defined in (\ref{eq:secants}).
The proof of Whitney's theorem shows that the tangent vectors are what determine the value of the objective function $D_{\Sigma}(p)$ in (\ref{eq:distortion}) \cite{book:hirsch-1997}.
One expects the (approximate) tangent vectors to be among the shortest secant vectors prior to normalization.
In order to reduce the size of the set $\Sigma$ only the $20$ shortest secant vectors were retained for each data point on the manifold.
Duplicates are also removed so that only the secant $\sigma$ or $-\sigma$ was included in $\Sigma$.

Here we'll examine classification of the handwritten digits.
A test image is also projected into each reduced space.
In each of the reduced spaces the (approximate) nearest neighbors to the projected test image are found.
Each of these nearest neighbors corresponds to an image in the original embedding space.
Then the test image can be reconstructed by combinations of the original images corresponding to the nearest neighbors in the reduced space.
The simplest reconstruction is to take the mean of the original images.
A test image is classified by which of these maps gives the best reconstruction.
In a sense, this provides an indication of how much compression of the data can be achieved while still having faithful image reconstruction.

Now allow the model of the image manifolds to be such that each image is given by an arbitrary affine transformation of a hypothetical template digit.
So, for example, there is a canonical `$0$' and any image of a handwritten zero is modeled as an affine map of this canonical image.
Then the dimensionality of the image manifolds would be six.
Whitney's theorem then gives $13$ as an appropriate embedding dimension.
In practice, relaxing the embedding dimension a little can help to retain structural information about a data manifold.
Here the manifolds were projected into \Rn{16} which still gives a $\sim\!98\%$ reduction in dimensionality versus the original images.
We used $15$ for the number of (approximate) nearest neighbors and took the mean of the corresponding original images as the estimate for a test image.
On the validation set the algorithm was robust to changes in the embedding dimension and number of nearest neighbors used.

On the test set the error rate was $2.80\%$ which compares favorably with the previous classification methods \cite{art:lecun-1998}\footnote{See \href{http://yann.lecun.com/exdb/mnist/}{http://yann.lecun.com/exdb/mnist/} for additional comparisons.}.
While not examined here, there are various enhancements that can be done for the classification.
The most obvious is to perturb the original images by general affine transformations and use these new images to `fill in' the data manifolds.
These transformations could be done on the initial nearest neighbors found above in order to enhance the reconstruction.
The original images could be combined using a weighted linear scheme where closer neighbors in the reduced space are weighted higher in the reconstruction.
Along this line, a quadratic surface could be fit to the original training data images and then that surface used to give the reconstructed test image.

The point of the Whitney embedding is to reduce the dimensionality of data, not to classify images.
What we've seen is that not much information about class membership is lost during this compression.
Given the $\sim\!98\%$ reduction in dimensionality, the modified Whitney embedding algorithm can be an effective preprocessing step to further classification algorithms beyond the simple linear reconstruction method used here.

\section{\label{sec:discussion}Discussion}
Reducing the dimensionality of a data set is a common preprocessing step.
For manifold-valued data an upper bound on the minimum embedding dimension is provided by Whitney's theorem.
This bound comes with the guarantee that the manifold structure of the data is maintained.
However, there can be significant distortion of the data during the projection into the lower-dimensional space.
And this distortion can vary over the manifold, being greater in some neighborhoods than others.

The procedure developed here attempts to minimize the overall distortion of the data.
Additionally, any two neighborhoods will tend to have more similar distortion which is a property of the function defined in (\ref{eq:distortion}).
Another feature of the current algorithm is that the embedding dimension of the data can be iteratively increased or decreased until a desired level of fidelity is attained.
In this case the Stiefel manifold, which maintains ordering of basis vectors, is a more natural setting than the Grassmannian.
Details for optimization over the Stiefel manifolds (i.e, formulas for the tangent spaces, metric and exponential map to replace those in (\ref{eq:Gnk-opt-info}) for the Grassmannian) are provided in \cite{misc:dreisigmeyer-2017,art:edelman-1999}.

Equation~(\ref{eq:opt-prob}) can be extended by replacing the $p \in \Gnk{m}{k}$ requirement with $p \in \Rn{m \times k}$.
For a given matrix $p$ the polar decomposition is $p = UP$ with $U$ unitary and $P$ symmetric positive-definite, $p$ assumed full-rank.
What is occurring in this situation is an initial projection $U$ followed by a stretching $P$ of the reduced-dimension embedding space.
The stretching undoes some of the distortion caused by the projection.

\bibliographystyle{siamplain}
\bibliography{references}
\end{document}